\documentclass[11pt]{article}
\usepackage{eacl2017}
\usepackage{times}
\usepackage{url}
\usepackage{latexsym}

\eaclfinalcopy % Uncomment this line for the final submission
\usepackage{pgfplots}
\pgfplotsset{compat=default} 

\usepackage{maltese}
\usepackage[utf8x]{inputenc}
\newcommand{\lemma}{\textsc}
\newcommand{\gherq}[1]{\lemma{√#1}}
\newcommand{\mt}[1]{\emph{#1}} %Maltese word alone
\newcommand{\en}[1]{`#1'} % Eng translation
\newcommand{\mten}[2]{\mt{#1} \en{#2}} % Maltese word with Eng translation
\newcommand{\mtgram}[2]{\mt{#1}, #2}
\newcommand{\naive}{Na\"{i}ve }

\title{Morphological Analysis for the Maltese Language:\\The challenges of a hybrid system}

\author{Claudia Borg \\
  Dept. A.I., Faculty of ICT \\
  University of Malta \\
  {\tt claudia.borg@um.edu.mt} \\\And
  Albert Gatt \\
  Institute of Linguistics \\
  University of Malta \\
  {\tt albert.gatt@um.edu.mt} \\}

\date{}

\begin{document}
\maketitle

\begin{abstract}
Maltese is a morphologically rich language with a hybrid morphological system which features both concatenative and non-concatenative processes. This paper analyses the impact of this hybridity on the performance of machine learning techniques for morphological labelling and clustering. In particular, we analyse a dataset of morphologically related word clusters to evaluate the difference in results for concatenative and non-concatenative clusters. We also describe research carried out in morphological labelling, with a particular focus on the verb category. Two evaluations were carried out, one using an unseen dataset, and another one using a gold standard dataset which was manually labelled. The gold standard dataset was split into concatenative and non-concatenative to analyse the difference in results between the two morphological systems.
\end{abstract}

\section{Introduction}\label{intro}

%%%% WANLP PAPER %%%%%%

Maltese, the national language of the Maltese Islands and, since 2004, also an official European language, has a hybrid morphological system that evolved from an Arabic stratum, a Romance (Sicilian/Italian) superstratum and an English adstratum \cite{Brincat:11}.  The Semitic influence is evident in the basic syntactic structure, with a highly productive non-Semitic component manifest in its lexis and morphology \cite{Fabri:10,Borg:97,Fabri:14}. Semitic morphological processes still account for a sizeable proportion of the lexicon and follow a non-concatenative, root-and-pattern strategy (or templatic morphology) similar to Arabic and Hebrew, with consonantal roots combined with a vowel melody and patterns to derive forms. By contrast, the Romance/English morphological component is concatenative (i.e. exclusively stem-and-affix based). Table \ref{1t_der_inf} provides an example of these two systems, showing inflection and derivation for the words \mten{eżamina}{to examine} taking a stem-based form, and \mten{gideb}{to lie} from the root \gherq{gdb} which is based on a templatic system. Table \ref{verbex} gives an examply of verbal inflection, which is affix-based, and applies to lexemes arising from both concatenative and non-concatenative systems, the main difference being that the latter evinces frequent stem variation.

\begin{table}[htp]
\caption[Examples of inflection and derivation in the two systems]{Examples of inflection and derivation in the concatenative and non-concatenative systems}
\begin{center}
\begin{tabular}{l|l|l}
 &  \textbf{Derivation} & \textbf{Inflection}\\
\hline
\textbf{Concat.} & & \\
 \mt{eżamina} & \mt{eżaminatur} & \mtgram{eżaminatr-iċi}{sg.f} \\
\en{examine} & \en{examiner} & \mtgram{eżaminatur-i}{pl.}\\
\hline
\textbf{Non-Con.} & & \\
\mten{gideb}{lie}  & \mt{giddieb} & \mtgram{giddieb-a}{sg.f.} \\
\gherq{gdb} & \en{liar}&  \mtgram{giddib-in}{pl.}\\
\hline
\end{tabular}
\end{center}
\label{1t_der_inf}
\end{table}%

\begin{table}[htbp]
\caption{Verbal inflections for the concatenative and non-concatenative systems.}
\begin{center}
\begin{tabular}{|c|c|c|}
\hline
 & \mt{eżamina} & \mt{gideb} \gherq{gdb} \\
 & \en{examine} &  \en{lie}  \\
\hline
\lemma{1Sg} & n-eżamina & n-igdeb \\
\lemma{2Sg} & t-eżamina & t-igdeb \\
\lemma{3SgM} & j-eżamina & j-igdeb \\
\lemma{3SgF}  & t-eżamina & t-igdeb\\
\lemma{1Pl}  & n-eżamina-w & n-igdb-u\\
\lemma{2Pl} & t-eżamina-w & t-igdb-u\\
\lemma{3Pl} & j-eżamina-w & j-igdb-u\\
\hline
\end{tabular}
\end{center}
\label{verbex}
\end{table}%

To date, there still is no complete morphological analyser for Maltese. In a first attempt at a computational treatment of Maltese morphology, \newcite{Farrugia:08b} used a neural network and focused solely on broken plural for nouns \cite{Schembri:06}. The only work treating computational morphology for Maltese in general was by \newcite{Borg:14}, who used unsupervised techniques to group together morphologically related words. A theoretical analysis of the templatic verbs \cite{Spagnol:11} was used by \newcite{Camilleri:13}, who created a computational grammar for Maltese for the Resource Grammar Library \cite{Ranta:11}, with a particular focus on inflectional verbal morphology. The grammar produced the full paradigm of a verb on the basis of its root, which can consist of over 1,400 inflective forms per derived verbal form, of which traditional grammars usually list 10. This resource is known as Ġabra and is available online\footnote{\url{http://mlrs.research.um.edu.mt/resources/gabra/}}. Ġabra is, to date, the best computational resource available in terms of morphological information. It is limited in its focus to templatic morphology and restricted to the wordforms available in the database. A further resource is the lexicon and analyser provided as part of the Apertium open-source machine translation toolkit \cite{apertium}. A subset of this lexicon has since been incorporated in the Ġabra database.

This paper presents work carried out for Maltese morphology, with a particular emphasis on the problem of hybridity in the morphological system. Morphological analysis is challenging for a language like Maltese due to the mixed morphological processes existing side by side. Although there are similarities between the two systems, as seen in verbal inflections, various differences among the subsystems exist which make a unified treatment challenging, including: (a) stem allomorphy, which occurs far more frequently with Semitic stems; (b) paradigmatic gaps, especially in the derivational system based on semitic roots \cite{Spagnol:11}; (c) the fact that morphological analysis for a hybrid system needs to pay attention to both stem-internal (templatic) processes, and phenomena occurring at the stem's edge (by affixation).

First, we will analyse the results of the unsupervised clustering technique by \newcite{Borg:14} applied on Maltese, with a particular focus of distinguishing the performance of the technique on the two different morphological systems. Second, we are interested in labelling words with their morphological properties. We view this as a classification problem, and treat complex morphological properties as separate features which can be classified in an optimal sequence to provide a final complex label.  Once again, the focus of the analysis is on the hybridity of the language and whether a single technique is appropriate for a mixed morphology such as that found in Maltese.

\section{Related Work}\label{background}

Computational morphology can be viewed as having three separate subtasks --- segmentation, clustering related words, and labelling (see \newcite{Hammarstrom:11}). Various approaches are used for each of the tasks, ranging from rule-based techniques, such as finite state transducers for Arabic morphological analysis \cite{Beesley:96,Habash:05b}, to various unsupervised, semi- or fully-supervised techniques which would generally deal with one or two of the subtasks. For most of the techniques described, it is difficult to directly compare results due to difference in the data used and the evaluation setting itself.  For instance, the results achieved by segmentation techniques are then evaluated in an information retrieval task.

The majority of works dealing with unsupervised morphology focus on English and assume that the morphological processes are concatenative \cite{Hammarstrom:11}. \newcite{Goldsmith:01} uses the minimum description length algorithm, which aims to represent a language in the most compact way possible by grouping together words that take on the same set of suffixes. In a similar vein, Creutz and Lagus \shortcite{Creutz:05a,Creutz:07} use Maximum a Posteriori approaches to segment words from unannotated texts, and have become part of the baseline and standard evaluation in the Morpho Challenge series of competitions \cite{Kurimo:10}. \newcite{Kohonen:10} extends this work by introducing semi- and supervised approaches to the model learning for segmentation. This is done by introducing a discriminative weighting scheme that gives preference to the segmentations within the labelled data. 

Transitional probabilities are used to determine potential word boundaries \cite{Keshava:06,Dasgupta:07,Demberg:07}. The technique is very intuitive, and posits that the most likely place for a segmentation to take place is at nodes in the trie with a large branching factor. The result is a ranked list of affixes which can then be used to segment words. 

\newcite{vandenBosch:99} and Clark \shortcite{Clark:02,Clark:07} apply Memory-based Learning to classify morphological labels. The latter work was tested on Arabic singular and broken plural pairs, with the algorithm learning how to associate an inflected form with its base form. \newcite{Durrett:13} derives rules on the basis of the orthographic changes that take place in an inflection table (containing a paradigm). A log-linear model is then used to place a conditional distribution over all valid rules. 

\newcite{Poon:09} use a log-linear model for unsupervised morphological segmentation, which leverages overlapping features such as morphemes and their context. It incorporates exponential priors as a way of describing a language in an efficient and compact manner. \newcite{Sirts:13} proposed Adaptor Grammars (AGMorph), a nonparametric Bayesian modelling framework for minimally supervised learning of morphological segmentation. The model learns latent tree structures over the input of a corpus of strings. \newcite{Narasimhan:15} also use a log-linear model, and morpheme and word-level features to predict morphological chains, improving upon the techniques of \newcite{Poon:09} and \newcite{Sirts:13}. A morphological chain is seen as a sequence of words that starts from the base word, and at each level through the process of affixation a new word is derived as a morphological variant, with the top 100 chains having an accuracy of 43\%. It was also tested on an Arabic dataset, achieving an F-Measure of 0.799. However, the system does not handle stem variation since the pairing of words is done on the basis of the same orthographic stem and therefore the result for Arabic is rather surprising. The technique is also lightly-supervised since it incorporates part-of-speech category to reinforce potential segmentations. 

Schone and Jurafsky \shortcite{Schone:00,Schone:01} and \newcite{Baroni:02} use both orthographic and semantic similarity to detect morphologically related word pairs, arguing that neither is sufficient on its own to determine a morphological relation. \newcite{Yarowsky:00} use a combination of alignment models with the aim of pairing inflected words. However this technique relies on part-of-speech, affix and stem information. \newcite{Can:12} create a hierarchical clustering of morphologically related words using both affixes and stems to combine words in the same clusters. \newcite{Ahlberg:14} produce inflection tables by obtaining generalisations over a small number of samples through a semi-supervised approach. The system takes a group of words and assumes that the similar elements that are shared by the different forms can be generalised over and are irrelevant for the inflection process. 

For Semitic languages, a central issue in computational morphology is disambiguation between multiple possible analyses. \newcite{Habash:05} learn classifiers to identify different morphological features, used specifically to improve part-of-speech tagging. \newcite{Snyder:08} tackle morphological segmentation for multiple languages in the Semitic family and English by creating a model that maps frequently occurring morphemes in different languages into a single abstract morpheme. 

Due to the intrinsic differences in the problem of computational morphology between Semitic and English/Romance languages, it is difficult to directly compare results. Our interest in the present paper is more in the types of approaches taken, and particularly, in seeing morphological labelling as a classification problem. Modelling different classifiers for specific morphological properties can be the appropriate approach for Maltese, since it allows the flexibility to focus on those properties where data is available.

\section{Clustering words in a hybrid morphological system}

The Maltese morphology system includes two systems, concatenative and non-concatenative. As seen in the previous section, most computational approaches deal with either Semitic morphology (as one would for Arabic or its varieties), or with a system based on stems and affixes (as in Italian). Therefore, we might expect that certain methods will perform differently depending on which component we look at. Indeed, overall accuracy figures may mask interesting differences among the different components.

The main motivation behind this analysis is that Maltese words of Semitic origin tend to have considerable stem variation (non-concatenative), whilst the word formation from Romance/English origin words would generally leave stems whole (concatenative)\footnote{Concatenative word formations would always involve a recognisable stem, though in some cases they may undergo minor variations as a result of allomorphy or allophomy.}. Maltese provides an ideal scenario for this type of analysis due to its mixed morphology. Often, clustering techniques would either be sensitive to a particular language, such as catering for weak consonants in Arabic \cite{DeRoeck:00}, or focus solely on English or Romance languages \cite{Schone:01,Yarowsky:00,Baroni:02} where stem variation is not widespread.

The analysis below uses a dataset of clusters produced by \newcite{Borg:14}, who employed an unsupervised technique using several interim steps to cluster words together. First, potential affixes are identified using transitional probabilities in a similar fashion to \cite{Keshava:06,Dasgupta:07}. Words are then clustered on the basis of common stems. Clusters are improved using measures of orthographic and semantic similarity, in a similar vein to \cite{Schone:01,Baroni:02}. Since no gold-standard lexical resource was available for Maltese, the authors evaluated the clusters using a crowd-sourcing strategy of non-expert native speakers and a separate, but smaller, set of clusters were evaluated using an expert group. In the evaluation, participants were presented with a cluster which had to be rated for its quality and corrected by removing any words which do not belong to a cluster. In this analysis, we focus on the experts' cluster dataset which was roughly balanced between non-concatenative (NC) and concatenative (CON) clusters. There are 101 clusters in this dataset, 25 of which were evaluated by all 3 experts, and the remaining by one of the experts. Table \ref{3t_clusterspread} provides an overview of the 101 clusters in terms of their size. 

\begin{table}[!htbp]
\caption{Comparison of non-concatenative and concatenative clusters in expert group}
\begin{center}
\begin{tabular}{|l|c|c|}
\hline
Size	& NC & CON\\
\hline
\textless 10	& 53\% (25)	& 26\% (14)\\
10--19	& 23\% (11) & 37\% (20)\\
20--29	& 13\% (6)	& 15\% (8)\\ 
30--39	& 2\% (1)	& 9\% (5) \\
\textgreater 40	& 9\% (4)	& 13\% (7) \\
\hline
Total	& 47 & 53 \\
Evaluated by all experts & 13 & 13\\
Evaluated by one expert  & 34 & 40\\
\hline
\end{tabular}
\end{center}
\label{3t_clusterspread}
\end{table}%

Immediately, it is possible to observe that concatenative clusters tend to be larger in size than non-concatenative clusters. This is mainly due to the issue of stem variation in the non-concatenative group, which gives rise to a lot of false negatives. It is also worth noting that part of the difficulty here is that the vowel patterns in the non-concatenative process are unpredictable. For example \mten{qsim}{division} is formed from \mten{qasam}{to divide} \gherq{qsm}, whilst \mten{ksur}{breakage} is formed from \mten{kiser}{to break} \gherq{ksr}. Words are constructed around infixation of vowel melodies to form a stem, before inflection adds affixes. In the concatenative system there are some cases of allomorphy, but there will, in general, be an entire stem, or substring thereof, that is recognisable.

\subsection{Words removed from clusters}

As an indicator of the quality of a cluster, the analysis looks at the number of words that experts removed from a cluster --- indicating that the word does not belong to a cluster. Table \ref{3t_wordperremovedspread} gives the percentage of words removed from clusters, divided according to whether the morphological system involved is concatenative or non-concatenative. The percentage of clusters which were left intact by the experts were higher for the concatenative group (61\%) when compared to the non-concatenative group (45\%).  The gap closes when considering the percentage of clusters which had a third or more of their words removed (non-concatenative at 25\% and concatenative at 20\%). However, the concatenative group also had clusters which had more than 80\% of their words removed. This indicates that, although in general the clustering technique performs better for the concatenative case, there are cases when bad clusters are formed through the techniques used. The reason is usually that stems with overlapping substrings are mistakenly grouped together. One such cluster was that for \mten{ittra}{letter}, which also got clustered with \mten{ittraduċi}{translate} and \mten{ittratat}{treated}, clearly all morphologically unrelated words. However, these were clustered together because the system incorrectly identified \mt{ittra} as a potential stem in all these words.

\begin{table}
\caption{Number of words removed, split by concatenative and non-concatenative processes}
\begin{center}
\begin{tabular}{|l|c|c|}
\hline
By Percentage & NC & CON	\\ \hline
0\% & 45\% (33) &  61\% (49)\\
1--5\% & 1\% (1)	&  1\% (1)	\\
5--10\%	& 7\% (5) &  4\% (3) \\
10--20\% & 5\% (4)	&  11\% (9)	\\
20--30\% & 17\% (12)	&  4\% (3)	\\
30--40\% & 8\% (6)	&  4\% (3)	\\
40--60\% & 7\% (5)	&  3\% (2)	\\
60--80\% & 10\% (7)	&  9\% (7)	\\
over 80\% & 0\% (0)	&  4\% (3)\\
\hline
\end{tabular}
\end{center}
\label{3t_wordperremovedspread}
\end{table}%

\subsection{Quality ratings of clusters}

Experts were asked to rate the quality of a cluster, and although this is a rather subjective opinion, the correlation between this judgement and the number of words removed was calculated using Pearson’s correlation coefficient. The trends are consistent with the analysis in the previous subsection; Table \ref{3t_qualityspread} provides the breakdown of the quality ratings for clusters split between the two processes and the correlation of the quality to the percentage of words removed. The non-concatenative clusters generally have lower quality ratings when compared to the concatenative clusters. But both groups have a strong correlation between the percentage of words removed and the quality rating, clearly indicating that the perception of a cluster's quality is related to the percentage of words removed.

\begin{table}[htbp]
\caption{Quality ratings of clusters, correlated to the percentage of words removed.}
\begin{center}
\begin{tabular}{|l|c|c|}
\hline
Quality	& NC & CON \\	 \hline
Very Good 	& 17\% (12) & 28\% (22) \\
Good 		& 33\% (24) & 36\% (29)	\\
Medium		& 34\% (25) & 18\% (15)	\\
Bad			& 12\% (9)	& 14\% (11)	\\
Very Bad	& 4\% (3)	& 4\% (3) \\
\hline \hline
Correlation: & 0.780 & 0.785 \\
\hline
\end{tabular}
\end{center}
\label{3t_qualityspread}
\end{table}%

\subsection{Hybridity in clustering}

Clearly, there is a notable difference between the clustering of words from concatenative and non-concatenative morphological processes. Both have their strengths and pitfalls, but neither of the two processes excel or stand out over the other. One of the problems with non-concatenative clusters was that of size. The initial clusters were formed on the basis of the stems, and due to stem variation the non-concatenative clusters were rather small. Although the merging process catered for clusters to be put together and form larger clusters, the process was limited to a maximum of two merging operations. This might not have been sufficient for the small-sized non-concatenative clusters. In fact, only 10\% of the NC clusters contained 30 or more words when compared to 22\% of the concatenative clusters. Limiting merging in this fashion may have resulted in a few missed opportunities. This is because there's likely to be a lot of derived forms which are difficult to cluster initially due to stem allomorphy (arising due to the fact that root-based derivation involves infixation, and in Maltese, vowel melodies are unpredictable). So there are possibly many clusters, all related to the same root. 

The problem of size with concatenative clusters was on the other side of the scale. Although the majority of clusters were of average size, large clusters tended to include many false positives. In order to explore this problem further, one possibility would be to check whether there is a correlation between the size of a cluster and the percentage of words removed from it. It is possible that the unsupervised technique does not perform well when producing larger clusters, and if such a correlation exists, it would be possible to set an empirically determined threshold for cluster size. 

Given the results achieved, it is realistic to state that the unsupervised clustering technique could be further improved using the evaluated clusters as a development set to better determine the thresholds in the metrics proposed above. This improvement would impact both concatenative and non-concatenative clusters equally. In general, the clustering technique does work slightly better for the concatenative clusters, and this is surely due to the clustering of words on the basis of their stems. This is reflected by the result that 61\% of the clusters had no words removed compared to 45\% of the non-concatenative clusters. However, a larger number of concatenative clusters had a large percentage of words removed. Indeed, if the quality ratings were considered as an indicator of how the technique performs on the non-concatenative vs the concatenative clusters, the judgement would be medium to good for the non-concatenative and good for the concatenative clusters. Thus the performance is sufficiently close in terms of quality of the two groups to suggest that a single unsupervised technique can be applied to Maltese, without differentiating between the morphological sub-systems.

\section{Classifying morphological properties}

In our approach, morphological labelling is viewed as a classification problem with each morphological property seen as a feature which can be classified. Thus, the analysis of a given word can be seen as a sequence of classification problems, each assigning a label to the word which reflects one of its morphological properties. We refer to such a sequence of classifiers as a `cascade'.

In this paper, we focus in particular on the verb category, which is morphologically one of the richest categories in Maltese. The main question is to identify whether there is a difference in the performance of the classification system when applied to lexemes formed through concatenative or non-concatenative processes. Our primary focus is on the classification of inflectional verb features. While these are affixed to the stem, the principal issue we are interested in is whether the co-training of the classifier sequence on an undifferentiated training set performs adequately on both lexemes derived via a templatic system and lexemes which have a `whole', continuous stem.

\subsection{The classification system}\label{class_sys}

The classification system was trained and initially evaluated using part of the annotated data from the lexical resource Ġabra. The training data contained over 170,000 wordforms, and the test data, which was completely unseen, contained around 20,000 wordforms.  A second dataset was also used which was taken from the Maltese national corpus (\lemma{mlrs} --- Malta Language Resource Server\footnote{\url{http://mlrs.research.um.edu.mt/}}). This dataset consisted of 200 randomly selected words which were given morphological labels by two experts. The words were split half and half between Semitic (non-concatenative) and Romance/English (concatenative) origin. The verb category had 94 words, with 76 non-concatenative, and 18 concatenative. This is referred to as the \textit{gold standard} dataset. 

A series of classifiers were trained using annotated data from Ġabra, which contains detailed morphological information relevant to each word. These are \textbf{person}, \textbf{number}, \textbf{gender}, \textbf{direct object}, \textbf{indirect object}, \textbf{tense}, \textbf{aspect}, \textbf{mood} and \textbf{polarity}. In the case of \textit{tense/aspect} and \textit{mood}, these were joined into one single feature, abbreviated to \textbf{TAM} since they are mutually exclusive. These features are referred to as \textit{second-tier} features, representing the morphological properties which the system must classify. The classification also relies on a set of \textit{basic} features which are automatically extracted from a given word. These are \textbf{stems}, \textbf{prefixes}, \textbf{suffixes} and \textbf{composite suffixes}, when available\footnote{Composite suffixes occur when more than one suffix is concatenated to the stem, usually with enclitic object and indirect object pronouns, as in \mten{qatil-hu-li}{he killed him for me}.}, \textbf{consonant-vowel patterns} and \textbf{gemination}.

A separate classifier was trained for each of the second-tier features. In order to arrive at the ideal sequence of classifiers, multiple sequences were tested and the best sequence identified on the basis of performance on held-out data (for more detail see \newcite{Borg:16}). Once the optimal sequence was established, the classification system used these classifiers as a cascade, each producing the appropriate label for a particular morphological property and passing on the information learnt to the following classifier. The verb cascade consisted of the optimal sequence of classifiers in the following sequence: Polarity (Pol), Indirect Object (Ind), Direct Object (Dir), Tense/Aspect/Mood (TAM), Number (Num), Gender (Gen) and Person (Per). 

The classifiers were trained using decision trees through the \lemma{weka} data mining software \cite{weka}, available both through a graphical user interface and as an open-source java library. Other techniques, such as Random Forests, SVMs and \naive Bayes, were also tested and produced very similar results. The classifiers were built using the training datasets. The first evaluation followed the traditional evaluation principles of machine learning, using the test dataset which contained unseen wordforms from Ġabra, amounting to just over 10\% of the training data. This is referred to as the \textit{traditional} evaluation. 

However, there are two main aspects in our scenario that encouraged us to go beyond the traditional evaluation. First, Ġabra is made of automatically generated wordforms, several of which are never attested (though they are possible) in the {\sc mlrs} corpus. Second, the corpus contains several other words which are not present in \mt{Ġabra}, especially concatenative word formations. Thus, we decided to carry out a gold standard (GS) evaluation to test the performance of the classification system on actual data from the {\sc mlrs} corpus. The evaluation in this paper is restricted to the verb category.  

\subsection{Evaluation Results}

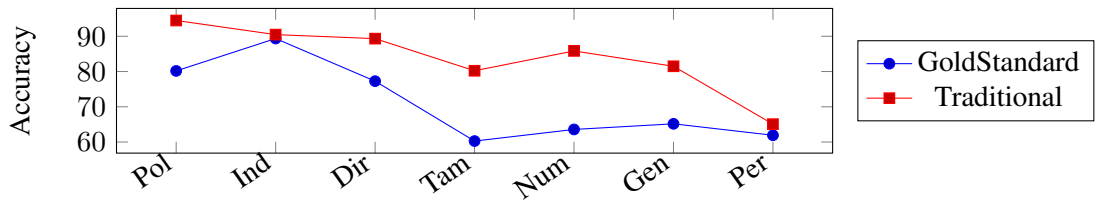
\begin{figure*}[!]
\begin{tikzpicture}
\begin{axis}[
	height=3.5cm,
	width=11cm,
    legend style={overlay, at={(1.2,0.5)}, anchor=center},
    ylabel={Accuracy},
    symbolic x coords={Pol,Ind,Dir,Tam,Num,Gen,Per},
    xtick=data,
    x tick label style={rotate=35,anchor=east},]
\addplot table[x = SemRom, y = PrevAvgACC] {pRomSemAcc.csv};
\addplot table[x = SemRom, y = Reg-Pred] {pRomSemAcc.csv};
\legend{GoldStandard,Traditional}
\end{axis}
\end{tikzpicture}
\caption{Comparison of the classification system using traditional evaluation settings and a gold standard evaluation.}
\label{pGS_Trad}
\end{figure*}

We first compare the performance of the classification system on the test dataset collected from \mt{Ġabra} to the manually annotated gold standard collated from the \lemma{mlrs} corpus. These results are shown in Figure \ref{pGS_Trad}. The first three features in the cascade --- Polarity, Indirect Object and Direct Object --- perform best in both the traditional and gold standard evaluations. In particular, the indirect object has practically the same performance in both evaluations. A closer look at the classification results of the words reveals that most words did not have this morphological property, and therefore no label was required. The classification system correctly classified these words with a {\em null} value. The polarity classifier on the other hand, was expected to perform better --- in Maltese, negation is indicated with the suffix \mt{-x} at the end of the word. The main problem here was that the classifier could apply the labels \textit{positive}, \textit{negative} or \textit{null} to a word, resulting in the use of the null label more frequently than the two human experts.
 
The errors in the classification of the morphological property TAM were mainly found in the labelling of the values \textit{perfective} and \textit{imperative}, whilst the label \textit{imperfective} performed slightly better. Similarly, the number and gender classifiers both had labels that performed better than others. Overall, this could indicate that the data representation for these particular labels is not adequate to facilitate the modelling of a classifier.

\begin{figure*}[h!tb]
\begin{tikzpicture}
\begin{axis}[
	height=3.5cm,
	width=11cm,
    legend style={overlay, at={(1.2,0.5)}, anchor=center},
    ylabel={Accuracy},
    symbolic x coords={Pol,Ind,Dir,Tam,Num,Gen,Per},
    xtick=data,
    x tick label style={rotate=35,anchor=east},]
\addplot table[x = SemRom, y = PrevAvgACC] {pRomSemAcc.csv};
\addplot table[x = SemRom, y = AvgAccROM] {pRomSemAcc.csv};
\addplot table[x = SemRom, y = AvgAccSEM] {pRomSemAcc.csv};
\legend{GoldStandard,Concatenative,Non-concatenative}
\end{axis}
\end{tikzpicture}
\caption{Comparison of the classifiers split between concatenative and non-concatenative words.}
\label{pGS_RomSem}
\end{figure*}
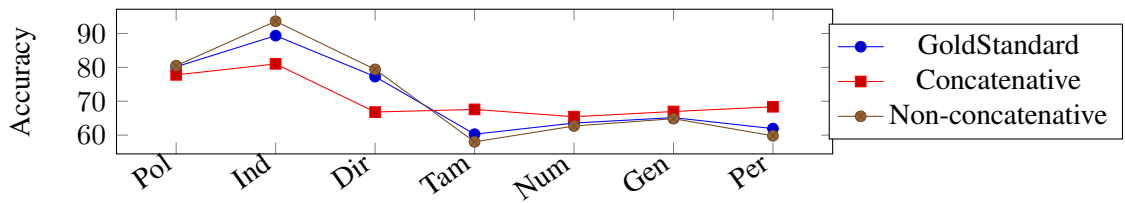

As expected, the performance of the classifiers on the gold standard is lower than that of a traditional evaluation setting. The test dataset used in the traditional evaluation, although completely unseen, was still from the same source as the training data (Ġabra) --- the segmentation of words was known, the distribution of instances in the different classes (labels) was similar to that found in the training data. While consistency in training and test data sources clearly make for better results, the outcomes also point to the possibility of overfitting, particularly as Ġabra contains a very high proportion of Semitic, compared to concatenative, stems. Thus, it is possible that the training data for the classifiers did not cover the necessary breadth for the verbs found in the \lemma{mlrs} corpus. To what extent this is impacting the results of the classifiers cannot be known unless the analysis separates the two processes. For this reason, the analysis of the verb category in the gold standard evaluation was separated into two, and the performance of each is compared to the overall gold standard performance. This allows us to identify those morphological properties which will require more representative datasets in order to improve their performance. Figure \ref{pGS_RomSem} shows this comparison.

The first three classifiers --- polarity, indirect object and direct object --- perform as expected, meaning that the concatenative lexemes perform worse than the non-concatenative. This confirms the suspicion that the coverage of Ġabra is not sufficiently representative of the morphological properties in the concatenative class of words. On the other hand, the TAM and Person classifiers perform better on the concatenative words. However, there is no specific distinction in the errors of these two classifiers.  

One overall possible reason for the discrepancy in the performance between the traditional and gold standard evaluation, and possibly also between the concatenative and non-concatenative words, is how the words are segmented. The test data in the traditional evaluation setting was segmented correctly, using the same technique applied for the training data. The segmentation for the words in the \lemma{mlrs} corpus was performed automatically and heuristically, and the results were not checked for their correctness, so the classification system might have been given an incorrect segmentation of a word. This would impact the results as the classifiers rely upon the identification of prefixes and suffixes to label words.

\section{Conclusions and Future Work}\label{conc}

This paper analysed the results of the clustering of morphologically related words and the morphological labelling of words, with a particular emphasis on identifying the difference in performance of the techniques used on words of Semitic origin (non-concatenative) and Romance/English origin (concatenative).  

The datasets obtained from the clustering technique were split into concatenative and non-concatenative sets, and evaluated in terms of their quality and the number of words removed from each cluster. Although generally, the clustering techniques performed best on the concatenative set, scalability seemed to be an issue, with the bigger clusters performing badly. The non-concatenative set, on the other hand, had smaller clusters but the quality ratings were generally lower than those of the concatenative group. Overall, it seems that the techniques were geared more towards the concatenative set, but performed at an acceptable level for the non-concatenative set. Although the analysis shows that it is difficult to find a one-size-fits-all solution, the resulting clusters could be used as a development set to optimise the clustering process in future. 

The research carried out in morphological labelling viewed it as a classification problem. Each morphological property is seen as a machine learning feature, and each feature is modelled as a classifier and placed in a cascade so as to provide the complete label to a given word. The research focussed on the verb category and two types of evaluations were carried out to test this classification system. The first was a traditional evaluation using unseen data from the same source as the training set. A second evaluation used randomly selected words from the \lemma{mlrs} corpus which were manually annotated with their morphological labels by two human experts. There is no complete morphological analyser available for Maltese, so this was treated as a gold standard. Since the classifiers were trained using data which is predominantly non-concatenative, the performance of the classification system on the \lemma{mlrs} corpus was, as expected, worse than the traditional evaluation. 

In comparing the two evaluations, it was possible to assess which morphological properties were not performing adequately. Moreover, the gold standard dataset was split into two, denoting concatenative and non-concatenative words, to further analyse whether a classification system that was trained predominantly on non-concatenative data could then be applied to concatenative data. The results were mixed, according to the different morphological properties, but overall, the evaluation was useful to determine where more representative data is needed. 

Although the accuracy of the morphological classification system are not exceptionally high for some of the morphological properties, the system performs well overall, and the individual classifiers can be retrained and improved as more representative data becomes available. And although the gold standard data is small in size, it allows us to identify which properties require more data, and of which type. One of the possible routes forward is to extend the grammar used to generate the wordforms in Ġabra and thus obtain more coverage for the concatenative process. However, it is already clear from the analysis carried out that the current approach is viable for both morphological systems and can be well suited for a hybrid system such as Maltese.

\section{Acknowledgements}

The authors acknowledge the insight and expertise of Prof. Ray Fabri. The research work disclosed in this publication is partially funded by the Malta Government Scholarship Scheme grant.

\bibliography{morphall}

\begin{thebibliography}{}

\bibitem[\protect\citename{Ahlberg \bgroup et al.\egroup }2014]{Ahlberg:14}
Malin Ahlberg, Markus Forsberg, and Mans Hulden.
\newblock 2014.
\newblock Semi-supervised learning of morphological paradigms and lexicons.
\newblock In {\em Proceedings of the 14th Conference of the European Chapter of
  the Association for Computational Linguistics, Gothenburg, Sweden 26--30
  April 2014}, pages 569--578.

\bibitem[\protect\citename{Baroni \bgroup et al.\egroup }2002]{Baroni:02}
Marco Baroni, Johannes Matiasek, and Harald Trost.
\newblock 2002.
\newblock Unsupervised discovery of morphologically related words based on
  orthographic and semantic similarity.
\newblock In {\em Proceedings of the ACL-02 workshop on Morphological and
  phonological learning - Volume 6}, MPL '02, pages 48--57. Association for
  Computational Linguistics.

\bibitem[\protect\citename{Beesley}1996]{Beesley:96}
Kenneth~R. Beesley.
\newblock 1996.
\newblock Arabic finite-state morphological analysis and generation.
\newblock In {\em Proceedings of the 16th conference on Computational
  linguistics}, pages 89--94. Association for Computational Linguistics.

\bibitem[\protect\citename{Borg and Azzopardi-Alexander}1997]{Borg:97}
Albert Borg and Marie Azzopardi-Alexander.
\newblock 1997.
\newblock {\em Maltese: Lingua Descriptive Grammar}.
\newblock Routledge, London and New York.

\bibitem[\protect\citename{Borg and Gatt}2014]{Borg:14}
Claudia Borg and Albert Gatt.
\newblock 2014.
\newblock Crowd-sourcing evaluation of automatically acquired, morphologically
  related word groupings.
\newblock In {\em Proceedings of the Ninth International Conference on Language
  Resources and Evaluation (LREC'14)}.

\bibitem[\protect\citename{Borg}2016]{Borg:16}
Claudia Borg.
\newblock 2016.
\newblock {\em {Morphology in the Maltese language: A computational
  perspective}}.
\newblock {Ph.D.} thesis, University of Malta.

\bibitem[\protect\citename{Brincat}2011]{Brincat:11}
Joseph~M. Brincat.
\newblock 2011.
\newblock {\em Maltese and other Languages}.
\newblock Midsea Books, Malta.

\bibitem[\protect\citename{Camilleri}2013]{Camilleri:13}
John~J. Camilleri.
\newblock 2013.
\newblock {A computational grammar and lexicon for Maltese}.
\newblock Master's thesis, Chalmers University of Technology, Gothenburg,
  Sweden, September.

\bibitem[\protect\citename{Can and Manandhar}2012]{Can:12}
Burcu Can and Suresh Manandhar.
\newblock 2012.
\newblock Probabilistic hierarchical clustering of morphological paradigms.
\newblock In {\em Proceedings of the 13th Conference of the European Chapter of
  the Association for Computational Linguistics}, pages 654--663. Association
  for Computational Linguistics.

\bibitem[\protect\citename{Clark}2002]{Clark:02}
Alexander Clark.
\newblock 2002.
\newblock Memory-based learning of morphology with stochastic transducers.
\newblock In {\em Proceedings of the 40th Annual Meeting of the Association for
  Computational Linguistics (ACL)}, pages 513--520.

\bibitem[\protect\citename{Clark}2007]{Clark:07}
Alexander Clark.
\newblock 2007.
\newblock {Supervised and Unsupervised Learning of Arabic Morphology}.
\newblock In Abdelhadi Soudi, Antal van~den Bosch, G\"unter Neumann, and Nancy
  Ide, editors, {\em Arabic Computational Morphology}, volume~38 of {\em Text,
  Speech and Language Technology}, pages 181--200. Springer Netherlands.

\bibitem[\protect\citename{Creutz and Lagus}2005]{Creutz:05a}
Mathias Creutz and Krista Lagus.
\newblock 2005.
\newblock Inducing the morphological lexicon of a natural language from
  unannotated text.
\newblock In {\em Proceedings of AKRR'05, International and Interdisciplinary
  Conference on Adaptive Knowledge Representation and Reasoning}, pages
  106--113.

\bibitem[\protect\citename{Creutz and Lagus}2007]{Creutz:07}
Mathias Creutz and Krista Lagus.
\newblock 2007.
\newblock Unsupervised models for morpheme segmentation and morphology
  learning.
\newblock {\em ACM Trans. Speech Lang. Process.}, 4(1):1--34.

\bibitem[\protect\citename{Dasgupta and Ng}2007]{Dasgupta:07}
Sajib Dasgupta and Vincent Ng.
\newblock 2007.
\newblock High-performance, language-independent morphological segmentation.
\newblock In {\em NAACL HLT 2007: Proceedings of the Main Conference}, pages
  155--163.

\bibitem[\protect\citename{de Roeck and Al-Fares}2000]{DeRoeck:00}
Anne~N. de~Roeck and Waleed Al-Fares.
\newblock 2000.
\newblock A morphologically sensitive clustering algorithm for identifying
  {A}rabic roots.
\newblock In {\em Proceedings of the 38th Annual Meeting on Association for
  Computational Linguistics}, ACL '00, pages 199--206. Association for
  Computational Linguistics.

\bibitem[\protect\citename{Demberg}2007]{Demberg:07}
Vera Demberg.
\newblock 2007.
\newblock A language-independent unsupervised model for morphological
  segmentation.
\newblock In {\em Proceedings of the 45th Annual Meeting of the Association of
  Computational Linguistics}, pages 920--927.

\bibitem[\protect\citename{Durrett and DeNero}2013]{Durrett:13}
Greg Durrett and John DeNero.
\newblock 2013.
\newblock Supervised learning of complete morphological paradigms.
\newblock In {\em Proceedings of the North American Chapter of the Association
  for Computational Linguistics}, pages 1185--1195.

\bibitem[\protect\citename{Fabri \bgroup et al.\egroup }2014]{Fabri:14}
Ray Fabri, Michael Gasser, Nizar Habash, George Kiraz, and Shuly Wintner.
\newblock 2014.
\newblock Linguistic introduction: The orthography, morphology and syntax of
  semitic languages.
\newblock In {\em Natural Language Processing of Semitic Languages}, Theory and
  Applications of Natural Language Processing, pages 3--41. Springer Berlin
  Heidelberg.

\bibitem[\protect\citename{Fabri}2010]{Fabri:10}
Ray Fabri.
\newblock 2010.
\newblock Maltese.
\newblock In Christian Delcourt and Piet van Sterkenburg, editors, {\em The
  Languages of the New EU Member States}, volume~88, pages 791--816. Revue
  Belge de Philologie et d'Histoire.

\bibitem[\protect\citename{Farrugia}2008]{Farrugia:08b}
Alex Farrugia.
\newblock 2008.
\newblock {A computational analysis of the Maltese broken plural}.
\newblock {Bachelor's Thesis, University of Malta}.

\bibitem[\protect\citename{Forcada \bgroup et al.\egroup }2011]{apertium}
M.~L. Forcada, M.~Ginest\'i-Rosell, J.~Nordfalk, J.~O'Regan, S.~Ortiz-Rojas,
  J.~A. P\'erez-Ortiz, F.~S\'anchez-Mart\'inez, G.~Ram\'irez-S\'anchez, and
  F.~M. Tyers.
\newblock 2011.
\newblock {Apertium}: a free/open-source platform for rule-based machine
  translation.
\newblock {\em Machine Translation}, 25(2):127--144.

\bibitem[\protect\citename{Goldsmith}2001]{Goldsmith:01}
John Goldsmith.
\newblock 2001.
\newblock Unsupervised learning of the morphology of a natural language.
\newblock {\em Computational Linguistics}, 27:153--198.

\bibitem[\protect\citename{Habash and Rambow}2005]{Habash:05}
Nizar Habash and Owen Rambow.
\newblock 2005.
\newblock Arabic tokenization, part-of-speech tagging and morphological
  disambiguation in one fell swoop.
\newblock In {\em Proceedings of the 43rd Annual Meeting on Association for
  Computational Linguistics}, ACL '05, pages 573--580, Stroudsburg, PA, USA.
  Association for Computational Linguistics.

\bibitem[\protect\citename{Habash \bgroup et al.\egroup }2005]{Habash:05b}
Nizar Habash, Owen Rambow, and George Kiraz.
\newblock 2005.
\newblock {MAGEAD: A Morphological Analyzer and Generator for the Arabic
  Dialects}.
\newblock In {\em Proceedings of the ACL Workshop on Computational Approaches
  to Semitic Languages}, pages 17--24. The Association for Computer
  Linguistics.

\bibitem[\protect\citename{Hall \bgroup et al.\egroup }2009]{weka}
Mark Hall, Eibe Frank, Geoffrey Holmes, Bernhard Pfahringer, Peter Reutemann,
  and Ian~H. Witten.
\newblock 2009.
\newblock The weka data mining software: an update.
\newblock {\em SIGKDD Explor. Newsl.}, 11(1):10--18.

\bibitem[\protect\citename{Hammarstr\"{o}m and Borin}2011]{Hammarstrom:11}
Harald Hammarstr\"{o}m and Lars Borin.
\newblock 2011.
\newblock Unsupervised learning of morphology.
\newblock {\em Computational Linguistics}, 37:309--350.

\bibitem[\protect\citename{Keshava and Pitler}2006]{Keshava:06}
Samarth Keshava and Emily Pitler.
\newblock 2006.
\newblock A simpler, intuitive approach to morpheme induction.
\newblock In {\em PASCAL Challenge Workshop on Unsupervised Segmentation of
  Words into Morphemes}, pages 31--35.

\bibitem[\protect\citename{Kohonen \bgroup et al.\egroup }2010]{Kohonen:10}
Oskar Kohonen, Sami Virpioja, and Krista Lagus.
\newblock 2010.
\newblock Semi-supervised learning of concatenative morphology.
\newblock In {\em Proceedings of the 11th Meeting of the ACL Special Interest
  Group on Computational Morphology and Phonology}, SIGMORPHON '10, pages
  78--86, Stroudsburg, PA, USA. Association for Computational Linguistics.

\bibitem[\protect\citename{Kurimo \bgroup et al.\egroup }2010]{Kurimo:10}
Mikko Kurimo, Sami Virpioja, Ville Turunen, and Krista Lagus.
\newblock 2010.
\newblock {Morpho Challenge competition 2005--2010: evaluations and results}.
\newblock In {\em Proceedings of the 11th Meeting of the ACL Special Interest
  Group on Computational Morphology and Phonology}, SIGMORPHON '10, pages
  87--95. Association for Computational Linguistics.

\bibitem[\protect\citename{Narasimhan \bgroup et al.\egroup
  }2015]{Narasimhan:15}
Karthik Narasimhan, Regina Barzilay, and Tommi~S. Jaakkola.
\newblock 2015.
\newblock An unsupervised method for uncovering morphological chains.
\newblock {\em Transactions of the Association for Computational Linguistics
  (TACL)}, 3:157--167.

\bibitem[\protect\citename{Poon \bgroup et al.\egroup }2009]{Poon:09}
Hoifung Poon, Colin Cherry, and Kristina Toutanova.
\newblock 2009.
\newblock Unsupervised morphological segmentation with log-linear models.
\newblock In {\em Proceedings of Human Language Technologies: The 2009 Annual
  Conference of the North American Chapter of the Association for Computational
  Linguistics}, NAACL '09, pages 209--217.

\bibitem[\protect\citename{Ranta}2011]{Ranta:11}
Aarne Ranta.
\newblock 2011.
\newblock {\em Grammatical framework: programming with multilingual grammars}.
\newblock CSLI studies in computational linguistics. CSLI Publications, Center
  for the Study of Language and Information, Stanford (Calif.).

\bibitem[\protect\citename{Schembri}2006]{Schembri:06}
Tamara Schembri.
\newblock 2006.
\newblock {The Broken Plural in Maltese: An Analysis}.
\newblock {Bachelor's Thesis, University of Malta}.

\bibitem[\protect\citename{Schone and Jurafsky}2000]{Schone:00}
Patrick Schone and Daniel Jurafsky.
\newblock 2000.
\newblock Knowledge-free induction of morphology using latent semantic
  analysis.
\newblock In {\em Proceedings of the 2nd workshop on Learning language in logic
  and the 4th conference on Computational natural language learning - Volume
  7}, ConLL '00, pages 67--72. Association for Computational Linguistics.

\bibitem[\protect\citename{Schone and Jurafsky}2001]{Schone:01}
Patrick Schone and Daniel Jurafsky.
\newblock 2001.
\newblock Knowledge-free induction of inflectional morphologies.
\newblock In {\em Proceedings of the second meeting of the North American
  Chapter of the Association for Computational Linguistics on Language
  technologies}, NAACL '01, pages 1--9. Association for Computational
  Linguistics.

\bibitem[\protect\citename{Sirts and Goldwater}2013]{Sirts:13}
Kairit Sirts and Sharon Goldwater.
\newblock 2013.
\newblock Minimally-supervised morphological segmentation using adaptor
  grammars.
\newblock {\em Transactions of the Association for Computational Linguistics},
  1:255--266.

\bibitem[\protect\citename{Snyder and Barzilay}2008]{Snyder:08}
Benjamin Snyder and Regina Barzilay.
\newblock 2008.
\newblock Unsupervised multilingual learning for morphological segmentation.
\newblock In {\em Proceedings of ACL-08: HLT}, pages 737--745, Columbus, Ohio.
  Association for Computational Linguistics.

\bibitem[\protect\citename{Spagnol}2011]{Spagnol:11}
Michael Spagnol.
\newblock 2011.
\newblock {\em {A tale of two morphologies. Verb structure and argument
  alternations in Maltese.}}
\newblock {Ph.D.} thesis, University of Konstanz.

\bibitem[\protect\citename{{Van den Bosch} and Daelemans}1999]{vandenBosch:99}
Antal {Van den Bosch} and Walter Daelemans.
\newblock 1999.
\newblock Memory-based morphological analysis.
\newblock In {\em Proceedings of the 37th annual meeting of the Association for
  Computational Linguistics on Computational Linguistics}, pages 285--292.

\bibitem[\protect\citename{Yarowsky and Wicentowski}2000]{Yarowsky:00}
David Yarowsky and Richard Wicentowski.
\newblock 2000.
\newblock Minimally supervised morphological analysis by multimodal alignment.
\newblock In {\em Proceedings of the 38th Annual Meeting on Association for
  Computational Linguistics}, ACL '00, pages 207--216, Stroudsburg, PA, USA.
  Association for Computational Linguistics.

\end{thebibliography}
\bibliographystyle{eacl2017}

\end{document}